\makeatletter\def\graphicscache@inhibit{true}\makeatother

\documentclass[letterpaper, 10 pt, conference]{IEEEtran}  %

\IEEEoverridecommandlockouts                              %

\usepackage[utf8]{inputenc}

\usepackage[hidelinks]{hyperref}
\usepackage[style=ieee,hyperref,natbib=true,backend=bibtex,firstinits,doi=false,%
     mincitenames=1,maxcitenames=2,maxbibnames=99,sorting=none,terseinits=false,hyperref=true]{biblatex}
\bibliography{references.bib}
\renewbibmacro*{bbx:savehash}{}%
\defbibheading{bibliography}[\bibname]{\section*{References}}

\usepackage{url}
\usepackage{graphicx}
\usepackage[usenames,dvipsnames]{xcolor}
\usepackage[draft,inline,nomargin]{fixme}
\fxsetup{theme=color}

\usepackage{amsmath}
\usepackage{amssymb}
\usepackage{bbm}
\usepackage{textcomp}
\usepackage{siunitx}

\usepackage{booktabs}
\usepackage{threeparttable}
\usepackage{multirow}

\usepackage[capitalize]{cleveref}

\usepackage{tikz}
\usepackage{tikz-3dplot}
\usepackage[normalem]{ulem}   
\usetikzlibrary{arrows}
\usetikzlibrary{positioning,calc}
\usetikzlibrary{decorations.pathreplacing}
\usetikzlibrary{decorations.markings}
\usetikzlibrary{fit}
\usetikzlibrary{shapes.callouts}
\usetikzlibrary{shapes.geometric}
\usetikzlibrary{matrix}
\usetikzlibrary{spy}
\usepackage{setspace}
\usepackage{csquotes}

\usepackage{pgfplots}
\pgfplotsset{compat=1.9}
\pgfplotsset{every tick label/.append style={font=\scriptsize}}
\usepgfplotslibrary{groupplots}
\usepgfplotslibrary{units}
\usepgfplotslibrary{colorbrewer}
\usepackage{pgfplotstable}

\usepackage{balance}
\usepackage{xfrac}

\usepackage{blindtext}

\usepackage{ifthen}
\usepackage{tabularx}
\usepackage{makecell}

\usepackage[export]{adjustbox}

\IfFileExists{graphicscache.sty}{\usepackage{graphicscache}}

\DeclareMathOperator*{\argmax}{arg\,max}

\DeclareMathOperator*{\mini}{min}
\usepackage{mathtools}
\usepackage{placeins}
\usepackage{dblfloatfix}
\newcommand{\rotateRPY}[3]%
{   \pgfmathsetmacro{\rollangle}{#1}
    \pgfmathsetmacro{\pitchangle}{#2}
    \pgfmathsetmacro{\yawangle}{#3}

    \pgfmathsetmacro{\newxx}{cos(\yawangle)*cos(\pitchangle)}
    \pgfmathsetmacro{\newxy}{sin(\yawangle)*cos(\pitchangle)}
    \pgfmathsetmacro{\newxz}{-sin(\pitchangle)}
    \path (\newxx,\newxy,\newxz);
    \pgfgetlastxy{\nxx}{\nxy};

    \pgfmathsetmacro{\newyx}{cos(\yawangle)*sin(\pitchangle)*sin(\rollangle)-sin(\yawangle)*cos(\rollangle)}
    \pgfmathsetmacro{\newyy}{sin(\yawangle)*sin(\pitchangle)*sin(\rollangle)+ cos(\yawangle)*cos(\rollangle)}
    \pgfmathsetmacro{\newyz}{cos(\pitchangle)*sin(\rollangle)}
    \path (\newyx,\newyy,\newyz);
    \pgfgetlastxy{\nyx}{\nyy};

    \pgfmathsetmacro{\newzx}{cos(\yawangle)*sin(\pitchangle)*cos(\rollangle)+ sin(\yawangle)*sin(\rollangle)}
    \pgfmathsetmacro{\newzy}{sin(\yawangle)*sin(\pitchangle)*cos(\rollangle)-cos(\yawangle)*sin(\rollangle)}
    \pgfmathsetmacro{\newzz}{cos(\pitchangle)*cos(\rollangle)}
    \path (\newzx,\newzy,\newzz);
    \pgfgetlastxy{\nzx}{\nzy};
}

\tikzset{RPY/.style={x={(\nxx,\nxy)},y={(\nyx,\nyy)},z={(\nzx,\nzy)}}}

\newcommand\givenbase[1][]{\,#1\lvert\,}
\let\given\givenbase

\DeclarePairedDelimiterX\Basics[1](){\let\given\sgiven #1}

\newcommand{\probP}{\text{I\kern-0.15em P}}
\newcommand{\STAB}[1]{\begin{tabular}{@{}c@{}}#1\end{tabular}}

\title{\LARGE \bf
  Learning Implicit Probability Distribution Functions for Symmetric Orientation Estimation from RGB Images Without Pose Labels
\thanks{$^{*}$ Equal contribution}
}

\author{
\IEEEauthorblockN{Arul Selvam Periyasamy$^{*}$}
\IEEEauthorblockA{\textit{Autonomous Intelligent Systems} \\
\textit{University of Bonn}\\
Bonn, Germany \\
Email: {\tt periyasa@ais.uni-bonn.de}}
\and
\IEEEauthorblockN{Luis Denninger$^{*}$}
\IEEEauthorblockA{\textit{Autonomous Intelligent Systems} \\
\textit{University of Bonn}\\
Bonn, Germany \\
Email: {\tt l\_denninger@uni-bonn.de}}
\and
\IEEEauthorblockN{Sven Behnke}
\IEEEauthorblockA{\textit{Autonomous Intelligent Systems} \\
\textit{University of Bonn}\\
Bonn, Germany \\
Email: {\tt behnke@cs.uni-bonn.de}}
}

\begin{document}

\maketitle

\begin{abstract}

Object pose estimation is a necessary prerequisite for autonomous robotic manipulation,
but the presence of symmetry increases the complexity of the pose estimation task.
Existing methods for object pose estimation output a single 6D pose. 
Thus, they lack the ability to reason about symmetries. 
Lately, modeling object orientation as a non-parametric probability distribution on the SO(3) manifold by neural networks has shown impressive results.
However, acquiring large-scale datasets to train pose estimation models remains a bottleneck.
To address this limitation, we introduce an automatic pose labeling scheme.
Given RGB-D images without object pose annotations and 3D object models, we design a two-stage pipeline consisting of point cloud registration and render-and-compare validation
to generate multiple symmetrical pseudo-ground-truth pose labels for each image.
Using the generated pose labels, we train an ImplicitPDF model to estimate the likelihood of an orientation hypothesis given an RGB image.
An efficient hierarchical sampling of the SO(3) manifold enables tractable generation of the complete set of symmetries at multiple resolutions.
During inference, the most likely orientation of the target object is estimated using gradient ascent.
We evaluate the proposed automatic pose labeling scheme and the ImplicitPDF model on a photorealistic dataset and the T-Less dataset,
demonstrating the advantages of the proposed method. 
\end{abstract}

\section{Introduction}
6D object pose estimation is the task of predicting the translation $\mathbf{t}\in\mathbb{R}^3$ and the orientation $\mathbf{R}\in\mathbf{SO}(3)$ of an object in the sensor coordinate frame.
It is a necessary prerequisite for autonomous robotic manipulation, industrial bin picking, as well as virtual and augmented reality.
With the advent of convolutional neural networks (CNNs), significant progress has been made in object estimation from RGB and RGB-D images.
We use the notation RGB-(D) to denote either RGB or RGB-D images.
Despite the improvements, large-scale object pose estimation remains challenging.
The challenges, particularly in orientation estimation, are compounded by object symmetries. 
Objects in our daily life and in industrial setups exhibit symmetries due to which it is impossible to estimate a single 6D pose only.
Ambiguities due to symmetries present hindrance in learning visual representations for pose estimation. 
Formally, ambiguities occur when an object $\mathcal{O}$ appears similar under at least two different poses $\mathbf{P}_i$ and $\mathbf{P}_j$,
i.e., we obtain the same image $\mathcal{I}$ when object $\mathcal{O}$ is in pose $\mathbf{P}_i$ and $\mathbf{P}_j$:

\begin{equation}
\label{eq:visSym}
\mathcal{I} \left( \mathcal{O}, \mathbf{P}_i \right) = \mathcal{I} \left( \mathcal{O}, \mathbf{P}_j \right).
\end{equation}

\begin{figure}
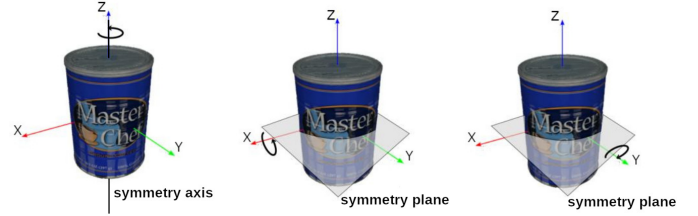

  \centering
   \setlength{\tabcolsep}{0.01cm}
   \newlength{\gtimgw}\setlength{\gtimgw}{2.5cm}
   \setlength{\gtimgw}{2.5cm}
  \begin{tabular}{ccc}
   \includegraphics[width=3cm]{figures/sym/symmetry_vis_1.png} &
   \includegraphics[width=3cm]{figures/sym/symmetry_vis_2.png} &
   \includegraphics[width=3cm]{figures/sym/symmetry_vis_3.png} \\
  \end{tabular}
    \caption{\textit{Can} object symmetries. Due to the presence of visual-symmetry-breaking texture, the \textit{can} object exhibits only geometric symmetries, namely 
    a continuous rotational symmetry along the  $z$ axis and a discrete flip symmetry along the $x$ and $y$ axes. 
    }
  \label{fig:can_sym}
\end{figure}

Symmetries can be classified into visual symmetries and geometric symmetries. 
Visual symmetries, as defined in \cref{eq:visSym}, arise due to the lack of distinctive visual features, whereas even in the presence of symmetry-breaking visual features,
objects may exhibit symmetries in terms of their geometry.
Knowing the object axes and type of symmetries, we can describe symmetries---despite the presence of textures---with respect 
to the object's geometry as discrete or continuous rotations around the object axes, as shown in ~\cref{fig:can_sym}.
Formally, an object $\mathcal{O}$ consisting of $n$ 3D points $\textbf{x}$  can be considered to exhibit geometric symmetries when 
there are at least two poses $\mathbf{P}_i$ and $\mathbf{P}_j$
that have a small mean closest point distance:

\begin{equation}\label{eqn:mesh_dist}
    \frac{1}{n}\sum_{\textbf{x}_1 \in \mathcal{O}}  \mini_{\textbf{x}_2 \in \mathcal{O}}|| \mathbf{P}_i\textbf{x}_1 - \mathbf{P}_j \textbf{x}_2 || \approx 0.
\end{equation}
Enumerating all sources of symmetries is not tractable, making an approach as shown in ~\cref{fig:can_sym} to describe arbitrary object symmetries not scalable.

Following the terminology introduced by \citet{bregier2018defining}, we define \textit{proper symmetries} $\mathcal{M}$ 
as the group of poses that exhibit geometric symmetries:
\begin{multline}\label{eqn:sym}
    \mathcal{M} = \{\mathbf{m} \in \mathbf{SE}(3) \; \text{such that} 
   \\
    \forall \mathbf{P} \in \mathbf{SE}(3),  
  \frac{1}{n}\sum_{\textbf{x}_1 \in \mathcal{O}}  \mini_{\textbf{x}_2 \in \mathcal{O}}|| \mathbf{P}_i\textbf{x}_1 - \mathbf{m} \cdot \mathbf{P}_j \textbf{x}_2 || \approx 0.\},
\end{multline}
with $\mathbf{m}$  being the transformations rotating the object around its symmetry axes in discrete steps.

The methods for pose estimation for symmetric objects can be classified into two families. 

The first family of methods estimates a single valid pose $\mathbf{p} \in \mathbf{SE}(3)$ corresponding to the RGB-(D) image.
One major advantage of this approach is that the CNNs architectures remain the same for symmetric and non-symmetric objects.
In both cases, given an RGB-(D) image, the network generates a single 6D pose prediction.
The only difference lies in the loss function used to train the model. Thus, in terms of the neural network architecture, training scheme,
and inference, both symmetric and non-symmetric objects  are treated the same.

The second family of methods estimates the complete set of \textit{proper symmetries} $\mathcal{M}$.
Modeling symmetries explicitly provides benefits that are twofold: i) facilitate models in learning better
visual representations and ii) better integration with downstream tasks.

Our work is based on the implicit probability distribution (ImplicitPDF) models for rotation manifolds introduced by~\citet{murphy2021implicit}.
Given an RGB image and a pose hypothesis, we train a CNN to estimate the likelihood of the pose hypothesis given the RGB image.
The novelty of our approach is that we do not need explicit ground-truth pose annotations to train the ImplicitPDF CNN model,
eliminating the bottleneck of acquiring large-scale ground-truth annotated dataset.

Given RGB-D images of symmetric objects without pose annotations, we propose an automatic pose labeling scheme and train the ImplicitPDF model using the generated pseudo ground-truth.
The model trained with the pseudo ground-truth is able to express the complete set of \textit{proper symmetries} $\mathcal{M}$ without prior knowledge about object symmetries.

\section{Related Work}

The state-of-the-art methods for predicting a single 6D pose are trained using ShapeMatch-Loss~\citep{xiang2018posecnn} for symmetric objects~\citep{hodavn2020bop, labbe2020, xu2022rnnpose, amini2022yolopose}.
Employing ShapeMatch-Loss implicitly selects one pose closest to the current pose prediction from the \textit{proper symmetry} set as ground-truth. 
While ShapeMatch-Loss does not need explicit definition of symmetry, during training, the ground-truth pose depends on the current pose prediction and this variability in ground-truth pose might hamper the models' learning ability.
In contrast, \citet{pitteri2019object, periyasamy2018pose} mapped the symmetrical rotations to a single \enquote{canonical} rotation.
One disadvantage of these methods is the requirement of explicit definition of object symmetry.
\citet{esteves2019cross} and \citet{saxena2009learning} proposed methods to learn features equivariant to specific symmetry classes.
In addition to learning pose estimation, \citet{rad2017bb8} added an auxiliary task to classify the type of symmetry an object exhibits. 
The authors argued that the auxiliary task helped the model to learn additional properties of the object's symmetry, which, in turn, benefit 6D pose estimation.
While the auxiliary task helped improving the single pose prediction accuracy, the formulation of the auxiliary task as a classification task limits its scope in modeling \textit{proper symmetries}.
\citet{corona2018pose} sidestepped the problem of estimating 6D pose and modeled pose estimation as a task of image comparison.
They trained a model to estimate a similarity score of two RGB images and used the similarity score to select the image from a codebook of images that
best matches the test image during inference. The pose corresponding to the matched image is considered as the pose of the target object in the test image.
In case of symmetric objects, multiple images from the codebook have a high similarity score. 
Since the inference involves comparing against a large-size codebook of images, the inference time requirement is high.
\citet{sundermeyer2020augmented} addressed this issue by using \textit{Augmented Autoencoder}, a variant of Denoising Autoencoder, to learn a low-dimensional
latent space representation for images and uses the latent space for image comparison. Despite the speed-up achieved in the codebook comparison, 
discretization of $\mathbf{SO}(3)$ is still needed to make the model real-time capable.
\citet{manhardt2019explaining} trained their model to generate a set of predictions given an RGB image with an intent to cover multiple possible poses.
In their experiments, they set the number of pose predictions to five. While their model can predict up to five correct poses in the presence of visual symmetry,
it is neither enough to cover the complete \textit{proper symmetries} nor does it reveal any information about the type of symmetry.
\citet{deng2022deep, gilitschenski2019deep} modeled multiple pose hypotheses as Bingham distributions and trained a CNN model to estimate the distribution parameters given an RGB-D input.
\citet{mohlin2020fisher} used Fisher distributions to model multiple pose hypotheses. These methods suffer under the complexity that arises in estimating the normalization constant of the distributions.
\citet{okorn2020learning} predicted a discrete histogram distribution of pose hypotheses by learning to predict the likelihood using comparison against a dictionary of images.
Our approach to model symmetries is inspired by ImplicitPDFs for rotational manifold introduced by \citet{murphy2021implicit}, which we discuss in~\cref{sec:ipdf} in detail. 
In contrast to \citet{murphy2021implicit}, we propose an automatic pose labeling scheme to generate multiple ground-truth annotations for each training image to train our ImplicitPDF model. 
\section{Method}

\begin{figure}
  \centering
    \includegraphics[width=\linewidth]{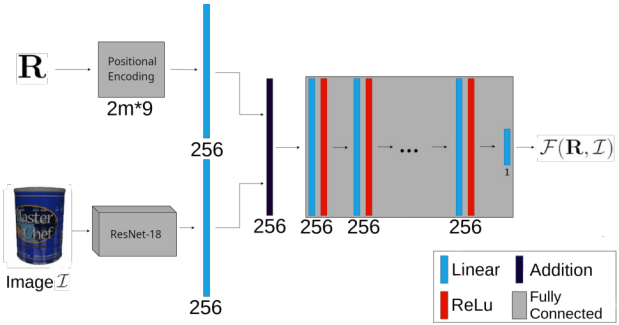}
    \caption{Learning ImplicitPDF. Given an image $\mathbf{I}$ and an orientation hypothesis $\mathbf{R}$, the CNN model is trained to generate the unnormalized joint log probability of the orientation hypothesis and the image.
    }
 \label{fig:model}
\end{figure}

\subsection{Implicit Probability Distribution for Object Orientation}
\label{sec:ipdf}
Inspired by the success of Neural Fields in modeling shape, scene, and rotation manifold~\citep{xie2022neural,mescheder2019occupancy,mildenhall2021nerf,murphy2021implicit}, 
we model object symmetries as a conditional probability distribution of the likelihood $\mathcal{P}(\mathbf{P} \given \mathcal{I})$ of the pose hypothesis $\mathbf{P}$ given an image $\mathcal{I}$ implicitly using a neural network.
To this end, we train a neural network $\mathcal{F}$ to predict the unnormalized joint log probability $\mathcal{F}(\mathbf{P}, \mathcal{I})$ of the pose hypothesis $\mathbf{P}$ and image $\mathcal{I}$ as shown in~\cref{fig:model}.
Let $\alpha$ be the normalization constant such that

\begin{equation}
    \mathcal{P}(\mathbf{P}, \mathcal{I}) =\alpha \exp ( \mathcal{F}(\mathbf{P}, \mathcal{I})).
\end{equation}

Using the product rule, 
\begin{equation}
    \mathcal{P}(\mathbf{P} \given \mathcal{I}) = \frac{\mathcal{P}(\mathbf{P}, \mathcal{I})}{\mathcal{P}(I)},
\end{equation}
where
\begin{equation}
    \label{eqn:marg}
 \mathcal{P}(I) =  \int_{\mathbf{P} \in \mathbf{SE}(3)} \mathcal{P}(\mathbf{P}, \mathcal{I}) d\mathcal{I}.
\end{equation}

For simplicity, we consider only the object orientation $\mathbf{R}\in\mathbf{SO}(3)$ instead of the 6D pose $\in\mathbf{SE}(3)$.
To make computing marginal probabilities tractable, we replace the continuous integral in~\cref{eqn:marg} with a discrete summation over a equivolumetric partitioning of $\mathbf{SO}(3)$ with $N$ partitions of
volume $V=\sfrac{\pi^2}{N}$,
and cancel out the normalization constant $\alpha$:
\begin{equation}
    \label{eqn:marg_dis}
    \mathcal{P}(\mathbf{R} \given \mathcal{I}) =  \frac{1}{V}\frac{\exp (\mathcal{F}(\mathbf{R}, \mathcal{I}))}{\sum_{i}^{N}   \exp (\mathcal{F}(\mathbf{R}_i, \mathcal{I}))}.
\end{equation}

For a detailed derivation of~\cref{eqn:marg_dis}, we refer to~\citep{murphy2021implicit}.

\subsubsection{Training}
\label{sec:training}
We train our model to minimize the negative log-likelihood of the ground-truth orientation $\mathbf{R}_{GT}$: 
\begin{equation}
    \mathcal{L}(\mathcal{I}, \mathbf{R}_{GT}) = -\text{log}(\mathcal{P}(\mathbf{R}_{GT} | \mathcal{I})).
\end{equation}
Following \cref{eqn:marg_dis}, we approximate the computation of the distribution $\mathcal{P}(\mathbf{R}_i | \mathcal{I})$ using $\mathbf{R}_i \in \{\mathbf{R}^0\}$, an equivolumetric grid covering $\mathbf{SO}(3)$ as in \cref{sec:sampling}. 
The orientation hypothesis $\mathbf{R}$ given to the model as input is represented using \textit{positional encoding}~\citep{transformer2017}.
\subsubsection{Inference}
\label{sec:inference}
During inference, given an image $\mathcal{I}$, we predict the (single) most plausible orientation $\mathbf{R}^{\ast}_{I}$ using gradient ascent starting from a set of initial hypotheses $\{\mathbf{R}^0\}$:
\begin{equation}
    \label{eqn:inf}
    \mathbf{R}^{\ast}_{I} = \argmax_{\mathbf{R}\in\mathbf{SO}(3)} \mathcal{F}(\mathbf{R}, \mathcal{I}).
\end{equation}

To generate the full distribution, we evaluate $\mathcal{P}(\mathbf{R}j \given \mathcal{I})$~:~$\mathbf{R}_j \in \{\mathbf{R}^n\}$ sampled equivolumetrically over $\mathbf{SO}(3)$.

\subsubsection{Equivolumetric Sampling and Visualization of $\mathbf{SO}(3)$}

\begin{figure}
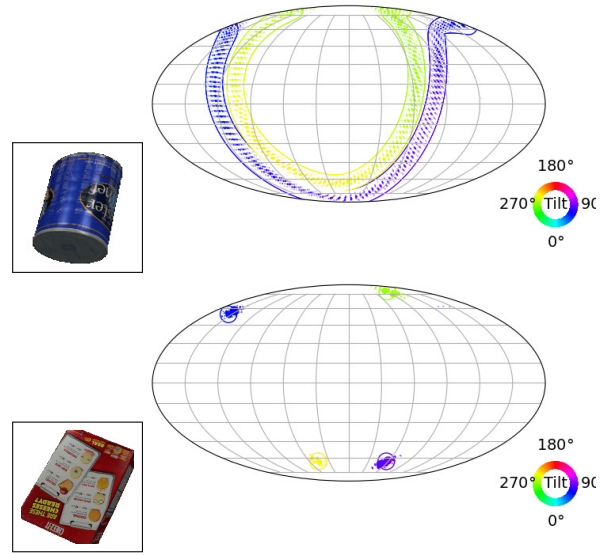

  \centering
 \setlength{\tabcolsep}{0.01cm}
 \newlength{\teaserimgw}\setlength{\teaserimgw}{2.5cm}
 \newlength{\inimgt}\setlength{\inimgt}{1.5cm}
 \newlength{\vizimgt}\setlength{\vizimgt}{6cm}
\begin{tabular}{cc}
    \framebox{\includegraphics[width=\inimgt, trim=50pt 50pt 50pt 50pt, clip]{figures/results/material_texture_big/image_0_0.png}} &
    \includegraphics[width=\vizimgt, trim=50pt 25pt 0pt 75pt, clip]{figures/results/material_texture_big/visualization_rot_0_0.png} \\
    \framebox{\includegraphics[width=\inimgt, trim=20pt 20pt 20pt 20pt, clip]{figures/results/box/image_raw_0_0.png}} &
 \includegraphics[width=\vizimgt, trim=50pt 25pt 0pt 75pt, clip]{figures/results/box/visualization_rot_0_0.png} \\
\end{tabular}
    \caption{Visualization of the ground-truth and the orientations predicted by the ImplicitPDF model for \textit{can} and \textit{box} objects.
    Given an RGB image and an orientation hypothesis, the ImplicitPDF model estimates the likelihood.
    The continuous lines and circles represent the ground-truth symmetries and the dots represent the orientation hypotheses with a high estimated likelihood.
    Two of three degrees of freedom of the SO(3) manifold are represented as a 2-sphere and projected on to a plane using Mollweide projection.
    The third degree of freedom is represented using a point on the color wheel.}
 \label{fig:teaser}
\end{figure}

\label{sec:sampling}
We follow the equivolumetric sampling of rotation manifold approach proposed by~\citet{murphy2021implicit} to generate $\{\mathbf{R}^0\}$ and $\{\mathbf{R}^n\}$ to cover $\mathbf{SO}(3)$ at different resolutions.
Using the HEALPix algorithm~\citep{healpix2005}  as a starting step, we generate equal area grids on the 2-sphere and iteratively use Hopf fibration~\citep{hopffibration}
to follow a great circle through each point on the surface of a 2-sphere to cover $\mathbf{SO}(3)$.
We also use the visualization method proposed by~\citet{murphy2021implicit} to visualize distributions of object orientations on the $\mathbf{SO}(3)$ manifold.
Rotation matrices in $\mathbf{SO}(3)$ have three degrees of freedom---two of the degrees of freedom are represented as a 2-sphere and projected on to a plane using Mollweide projection. The third degree of freedom is represented using Hopf fibration by a great circle of points  
to each point on the 2-sphere. The location of a point on the great circle is represented using a color wheel as shown in~\cref{fig:teaser}.
The number of samples generated in iteration $S_i$ is given by $72 \cdot 8^{S_i}$.

\subsubsection{Evaluation Metrics}
To evaluate the performance of the proposed method, we use the \textit{log-likelihood} (LLH) and the \textit{mean absolute angular error} (MAAD) metrics.
Given a set of ground-truth annotations $\mathbf{R}_{GT}$, 
the LLH metric measures the likelihood of the ground-truth orientations:
$$\text{LLH}(\mathbf{R}) =  \mathop{\mathbb{E}_{I\sim\mathcal{P}(I)}}\mathop{\mathbb{E}_{\mathbf{R}\sim\mathcal{P_{GT}}(\mathbf{R} | \mathcal{I})}} \text{log}(\mathcal{P}(\mathbf{R} | \mathcal{I})).$$
To compute log-likelihood, we do not need the complete set of \textit{proper symmetries}.
The standard \textit{mean absolute angular deviation}  (MAAD) is defined as:

$$\text{MAAD}(\mathbf{R}) =  \mathop{\mathbb{E}_{\mathbf{R}\sim\mathcal{P}(\mathbf{R} | \mathcal{I})}} \bigl[ \text{min}_{\mathbf{R}'\in{\mathbf{R}_{GT}}} d (R, R')  \bigr],$$
where $d$ is the geodesic distance between rotations.
                      
Furthermore, we report Recall MAAD as a measure of recall.
We extract a set of orientations $\{\mathbf{\hat{R}}\}$ with a predicted probability threshold of 1e-3. For each orientation in $\{\mathbf{R}_{GT}\}$, we 
find the closest orientation in $\{\mathbf{\hat{R}}\}$ in terms of the geodesic distance and report the mean of the shortest angular distance over the set $\{\mathbf{R}_{GT}\}$.
In the case of continuous symmetries, we discretize the symmetries into 200 orientations for computing the Recall MAAD metric.

\subsection{Learning Without Ground-Truth Annotations}

\begin{figure*}
   \centering
    \includegraphics[width=0.9\linewidth]{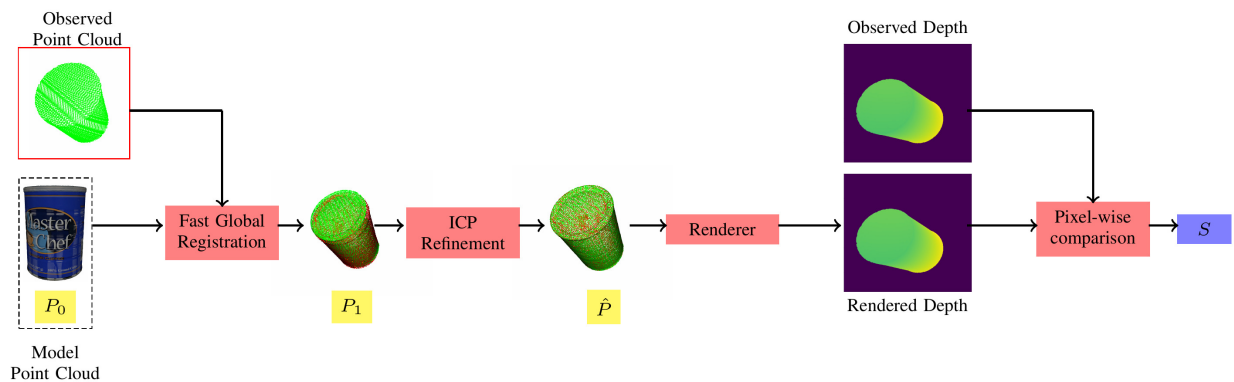}
    
    \vspace*{-2mm}
    \caption{Pseudo ground-truth generation process. Given an RGB-D image without pose annotation, we register the model point cloud in a random pose $P_0$
    against the observed point cloud using the Fast Global Registration algorithm to generate hypothesis $P_1$, which is
    further refined using the ICP algorithm utilizing the depth information to generate $\hat{P}$. We then render the model according to $\hat{P}$ to generate the depth image.
    The rendered depth image is compared with the observed depth image pixel-wise to compute the comparison score $S$. For a given RGB-D image we run this process multiple times
    and select the $\hat{P}$ with the smallest $S$.}
\label{fig:rnc}
\end{figure*}

\citet{murphy2021implicit} introduced the SYMSOL~I and SYMSOL~II datasets to benchmark symmetry learning methods. The datasets consist of renderings of platonic solids (tetrahedron, cube, icosahedron),
cone, and cylinder and corresponding ground-truth symmetries. In real-world applications, acquiring ground-truth symmetry annotations is prohibitively expensive. To address this issue, we propose a two-stage
automatic pose labeling scheme as illustrated in~\cref{fig:rnc}.

\begin{figure}
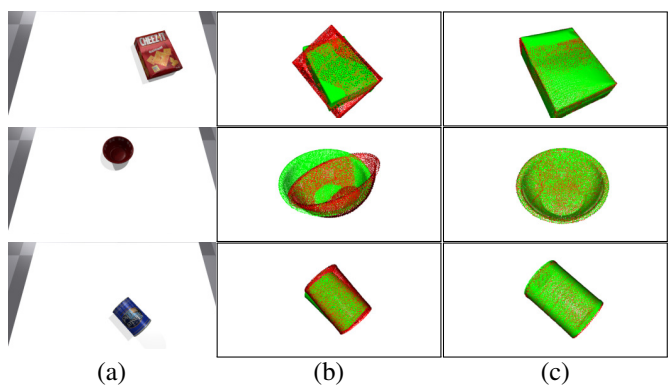

  \centering
 \setlength{\tabcolsep}{0.01cm}
 \setlength{\gtimgw}{2.75cm}
\begin{tabular}{ccc}
 \includegraphics[width=\gtimgw, trim=0pt 50pt 100pt 100pt, clip]{figures/gt_rnc/paper_vis_box/org_image.png} &
 \framebox{\includegraphics[width=\gtimgw, trim=0 50 0 40]{figures/gt_rnc/paper_vis_box/global_result.png}} &
    \framebox{\includegraphics[width=\gtimgw, trim=0 50 0 40]{figures/gt_rnc/paper_vis_box/local_result.png}} \\
 \includegraphics[width=\gtimgw, trim=0pt 50pt 100pt 100pt, clip]{figures/gt_rnc/paper_vis_bowl/org_image.png} &
 \framebox{\includegraphics[width=\gtimgw, trim=0 50 0 40]{figures/gt_rnc/paper_vis_bowl/global_result.png}} &
    \framebox{\includegraphics[width=\gtimgw, trim=0 50 0 40]{figures/gt_rnc/paper_vis_bowl/local_result.png}} \\
 \includegraphics[width=\gtimgw, trim=0pt 50pt 100pt 100pt, clip]{figures/gt_rnc/paper_vis_can/org_image.png} &
 \framebox{\includegraphics[width=\gtimgw, trim=0 50 0 40]{figures/gt_rnc/paper_vis_can/global_result.png}} &
    \framebox{\includegraphics[width=\gtimgw, trim=0 50 0 40]{figures/gt_rnc/paper_vis_can/local_result.png}} \\
    (a) & (b) & (c)\\
\end{tabular}
    \caption{Visualization of pseudo ground-truth generation. (a) RGB image of the scene. (b) Pose generated by the Fast Global Registration algorithm. (c) Final pseudo ground-truth pose generated by ICP optimization.
    In (b) and (c), the point cloud in ground-truth pose are visualized in green and the point cloud in generated pseudo ground-truth poses are visualized in red.}
 \label{fig:gt_gen}
\end{figure}

Given an RGB-D image of a scene and the 3D mesh of the target object, we start with unprojecting the depth map into 3D to generate the observed point cloud $C_{obs}$.
We transform the model point cloud $C_{model}$ according to a random initial pose $P_0$ and perform global registration of the transformed model point cloud against the observed point cloud
to generate a pose hypothesis $P_1$. We employ the \textit{Fast Global Registration} algorithm~\citep{fgr2016} in conjunction with \textit{Fast Point Feature Histogram} (FPFH) 
feature~\citep{fpfh2009} to perform global registration. 
We refine $P_1$ using the \textit{Iterative Closest Point} (ICP) algorithm to generate $\hat{P}$. The results of the different stages of the pipeline are visualized in \cref{fig:gt_gen}.
Transforming $C_{model}$ to a random initial pose $P_0$ at the beginning ensures that we generate a different $\hat{P}$ every time we run the process for an RGB-D image. This way, we generate
a set of pseudo ground-truth pose labels for each image in the training set.
The variability in the set of generated pseudo ground-truth pose labels for an image is vital for the model in learning the complete set of \textit{proper symmetries}.
Due to self-occlusion, an object is only partially visible in an RGB-D image. 
Without the knowledge of camera view direction, registering the complete model point cloud $C_{model}$ against the partial observed point cloud $C_{obs}$ might result in bad registrations
and it is not possible to detect the bad registrations based on the standard $\ell_2$ distance metric. To address this issue, we utilize the render-and-compare framework.
We render the depth map according to $\hat{P}$ and compare it pixel-wise with the observed depth map. In the ideal scenario, the render-and-compare difference should be close to zero.
To generate one pseudo ground-truth label, we repeat the process multiple times and select the $\hat{P}$ with the smallest comparison score.

\subsection{Dataset}

\begin{figure}
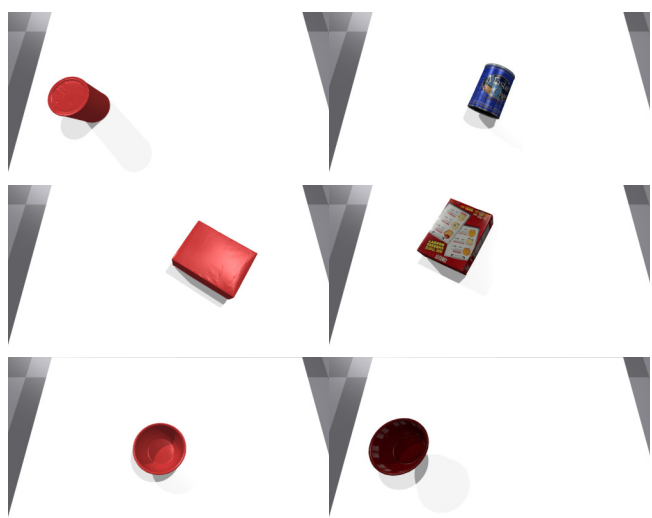

  \centering
 \setlength{\tabcolsep}{0.01cm}
 \newlength{\dsimgw}\setlength{\dsimgw}{4.25cm}
\begin{tabular}{cc}
 \includegraphics[width=\dsimgw, trim=0pt 50pt 100pt 100pt, clip]{figures/dataset/00_rgb.png} &
 \includegraphics[width=\dsimgw, trim=0pt 50pt 100pt 100pt, clip]{figures/dataset/rgb_can_mat.png} \\
 \includegraphics[width=\dsimgw, trim=0pt 50pt 100pt 100pt, clip]{figures/dataset/rgb_crackerbox_uni.png} &
 \includegraphics[width=\dsimgw, trim=0pt 50pt 100pt 100pt, clip]{figures/dataset/box_rgb.png} \\
 \includegraphics[width=\dsimgw, trim=0pt 50pt 100pt 100pt, clip]{figures/dataset/rgb_bowl_uni.png} &
 \includegraphics[width=\dsimgw, trim=0pt 50pt 100pt 100pt, clip]{figures/dataset/bowl_rgb.png} \\
\end{tabular}
    \caption{Exemplar RGB images from the Uniform and Texture dataset. First column: Uniform dataset. Second column: Texture dataset.}
 \label{fig:dataset}
\end{figure}

To evaluate the proposed method, we generate a photorealistic dataset using the Isaac GYM framework~\citep{nvidia2021isaac}.
The dataset consists of three objects---\textit{can}, \textit{box}, and \textit{bowl} (shown in~\cref{fig:dataset})---placed in randomly-sampled physically-plausible poses on a tabletop.
Each frame in the dataset consists of an RGB-D image, 6D object pose (single pose used to render the image) and segmentation ground-truth. 
Moreover, to evaluate the impact of the object texture over the model's ability to learn symmetries, we generate two versions of the dataset: using a uniform
colored texture and using the original material texture. We call these datasets {\em Uniform} and {\em Texture}, respectively. Both datasets consist of 20K images.
20\% of the samples in each dataset are used for validation and the rest are used for training.

\section{Experiments}
\begin{figure}
    \centering
		\resizebox{.99\linewidth}{!}{%
\begin{tikzpicture}
\begin{axis}[
    xlabel={\scriptsize{\# epochs} },
    ylabel={\scriptsize{Loss \& LLH}},
    xmin=1, xmax=30,
    ymin=-6, ymax=6,
    xtick={5,10,...,30},
    ytick={-6,-5.5,...,6},
    grid = both,
    legend columns=2,
legend style={
         nodes={scale=0.75},
         at={(0.5,-0.15)},anchor=north,
         font=\footnotesize
      }]
\addplot[smooth,color=red, densely dashed] plot coordinates {
(1,	1.53518867611292)
(2,	0.618102061288867)
(3,	0.103049117330443)
(4,	-0.232627141488651)
(6,	-0.529767211582234)
(7,	-0.780450925008574)
(8,	-1.02657043666982)
(9,	-1.2576390036303)
(10, -1.44236072912738)
(12, -1.62605027061197)
(13, -1.7651473022812)
(14, -1.90347968583083)
(15, -2.03328071542047)
(16, -2.1469981533971)
(18, -2.2523812518191)
(19, -2.34321140116127)
(20, -2.44975464498226)
(21, -2.51593607337914)
(22, -2.60583745069172)
(24, -2.67712964940427)
(25, -2.74711643759884)
(26, -2.83724452607074)
(27, -2.87840937856418)
(28, -2.95038491457849)
(30, -3.03330637092021)
(31, -3.06215505220404)
(32, -3.13409791301139)
(33, -3.16282638862951)
(34, -3.23212859642446)
(36, -3.26723007539018)
(37, -3.32288019336871)
(38, -3.37499321278055)
(39, -3.38392064820475)
(40, -3.43437611760192)
(42, -3.48437966517548)
(43, -3.5262041685)
(44, -3.53411197662354)
(45, -3.58375836367631)
(46, -3.64973296929355)
(48, -3.63414848028724)
(49, -3.68526695260954)
(50, -3.72836303473705)
(51, -3.74850152262408)
(52, -3.78584974084921)
(54, -3.79197715882638)
(55, -3.83859507598687)
(56, -3.87494839127384)
(57, -3.85389411627357)
(58, -3.91404186789669)
(60, -3.93415632888452)

};
\addlegendentry{Can Texture Training Loss}

\addplot[smooth, color=red]plot coordinates{
(1,	-1.27898081099831)
(2,	-0.864068417354888)
(3,	-0.820738695069641)
(4,	-0.687679329923368)
(6,	-0.432472074334074)
(7,	-0.279092428862563)
(8,	-0.089536905703246)
(9,	0.144880059061468)
(10, 0.35615515302185)
(12, 0.60412243281227)
(13, 0.732470636156168)
(14, 0.944853140352012)
(15, 0.990478314938123)
(16, 1.16244450956169)
(18, 1.20336403467534)
(19, 1.31401158803251)
(20, 1.44353615358347)
(21, 1.45656602154377)
(22, 1.54632421662595)
(24, 1.67905899909859)
(25, 1.63479628809821)
(26, 1.732706900611)
(27, 1.80707556364795)
(28, 1.92468806525652)
(30, 1.86578452468406)
(31, 1.98860291418656)
(32, 2.01051306238023)
(33, 2.13979327550896)
(34, 2.04014176794196)
(36, 2.04210192263161)
(37, 2.15687082864327)
(38, 2.13217043430168)
(39, 2.16477779131398)
(40, 2.18499028123366)
(42, 2.22548428783988)
(43, 2.15872136293641)
(44, 2.30780317949944)
(45, 2.265955111683)
(46, 2.38145562529177)
(48, 2.41788245111972)
(49, 2.43703319408416)
(50, 2.27393043135551)
(51, 2.3144800348433)
(52, 2.27604459780498)
(54, 2.36128112877514)
(55, 2.36689759984945)
(56, 2.32778798470128)
(57, 2.35201150728359)
(58, 2.37542216271356)
(60, 2.40584593974983)

};
\addlegendentry{Can Texture Log-Likelihood}

\addplot[smooth,color=blue, densely dashed] plot coordinates {
(1,	0.577038827338325)
(2,	-1.15390372709997)
(3,	-2.0055272395338)
(4,	-2.47145510490854)
(6,	-2.80886471805288)
(7,	-3.06428133907603)
(8,	-3.27679598509376)
(9,	-3.46110300875422)
(10, -3.60616778259847)
(12, -3.74069463554306)
(13, -3.86159952956053)
(14, -3.9732740316818)
(15, -4.06460332396019)
(16, -4.15047805107648)
(18, -4.23158281122274)
(19, -4.31143355725416)
(20, -4.38405707938161)
(21, -4.46559080792897)
(22, -4.52318112767158)
(24, -4.59532853501353)
(25, -4.65337474547808)
(26, -4.70719877641592)
(27, -4.76967596651903)
(28, -4.8177928473819)
(30, -4.86773125330607)
(31, -4.91590586467762)
(32, -4.97286771185956)
(33, -4.9910560057531)
(34, -5.04734257085999)
(36, -5.08592984213758)
(37, -5.15380929595795)
(38, -5.14724669290419)
(39, -5.20372650279335)
(40, -5.23237961678956)
(42, -5.27009640641473)
(43, -5.31520416250276)
(44, -5.32287217609918)
(45, -5.35838738009704)
(46, -5.39853806756622)
(48, -5.41797529761471)
(49, -5.42932509901512)
(50, -5.47857698279234)

};
\addlegendentry{Bowl Texture Training Loss}

\addplot[smooth, color=blue] plot coordinates {
(1,	0.2122736506023)
(2,	1.48484411400247)
(3,	2.07002759249172)
(4,	2.36351344086562)
(6,	2.5716386969398)
(7,	2.71307542365831)
(8,	2.85512332695128)
(9,	2.90350120251152)
(10, 2.93034048457829)
(12, 3.10283400192413)
(13, 3.1321249946932)
(14, 3.11174112808587)
(15, 3.06203117944402)
(16, 3.12491181848652)
(18, 3.06400952973124)
(19, 2.98812910256089)
(20, 3.04763153558619)
(21, 2.9692092705805)
(22, 2.9774243345143)
(24, 2.97291339759808)
(25, 2.85045748499989)
(26, 2.65380280671025)
(27, 2.79155411750006)
(28, 2.80026447709395)
(30, 2.60781398766462)
(31, 2.52803491093088)
(32, 2.59458570027962)
(33, 1.73028909193393)
(34, 1.96669166062956)
(36, 2.07564648767213)
(37, 1.61767052748369)
(38, 1.58331555770642)
(39, 1.50203046166399)
(40, 1.22184322446077)
(42, 0.83501788038416)
(43, 1.22725060160999)
(44, 0.474450387687945)
(45, 0.715175246128464)
(46, 0.596985381360613)
(48, 0.247515112265795)
(49, -0.138143866957625)
(50, -0.442884436875188)

};
\addlegendentry{Bowl Texture Log-Likelihood}

\addplot[smooth, color=violet, densely dashed] plot coordinates {
(1,	1.8802829269153)
(2,	1.21622586102035)
(3,	0.696171409455105)
(4,	0.27339841925831)
(6,	-0.12367324159476)
(7,	-0.490931540041865)
(8,	-0.838455156603856)
(9,	-1.1122116665935)
(10, -1.38291205043223)
(12, -1.61648279042979)
(13, -1.83970828198675)
(14, -2.04766046882269)
(15, -2.21254414290338)
(16, -2.37655305981043)
(18, -2.52943236258493)
(19, -2.6702239489674)
(20, -2.7946179458751)
(21, -2.89742081200899)
(22, -2.99349226999046)
(24, -3.08237187423516)
(25, -3.17489320010095)
(26, -3.2402071229261)
(27, -3.32032515990793)
(28, -3.39303877460423)
(30, -3.44746458115269)
(31, -3.50920355616517)
(32, -3.55671800546978)
(33, -3.59976618325532)
(34, -3.6601491828463)
(36, -3.6884617022614)
(37, -3.74734777953494)
(38, -3.76916073566646)
(39, -3.8242890751777)
(40, -3.86053123877416)
(42, -3.90555890401204)
(43, -3.93810552269665)
(44, -3.94672178984875)
(45, -4.00334938011359)
(46, -4.01751539837662)
(48, -4.0368671239312)
(49, -4.07253354461632)
(50, -4.08955433238205)
(51, -4.14246374339014)
(52, -4.14314019857948)
(54, -4.15586504058458)
(55, -4.18852104120587)
(56, -4.21509052271867)
(57, -4.24626213875576)
(58, -4.22189408985537)
(60, -4.2712347282106)

};
\addlegendentry{Box Texture Training Loss}

\addplot[smooth, color=violet] plot coordinates {
(1,	-1.70734446038414)
(2,	-1.01731702314524)
(3,	-0.564275202190145)
(4,	-0.126549976767103)
(6,	0.319955731480853)
(7,	0.656870823721055)
(8,	0.999801633329641)
(9,	1.35247372688626)
(10, 1.62068438802151)
(12, 1.93207967960174)
(13, 2.07132516250846)
(14, 2.38661760713476)
(15, 2.51276630138003)
(16, 2.73019388060042)
(18, 2.97422348190697)
(19, 2.99108551856791)
(20, 3.12532812794508)
(21, 3.25368469500452)
(22, 3.31288678605272)
(24, 3.45912004906475)
(25, 3.58166575763814)
(26, 3.56613213050473)
(27, 3.75585118714592)
(28, 3.795074096547)
(30, 3.88024427263108)
(31, 3.9302439903219)
(32, 3.90770245444361)
(33, 3.88866123074777)
(34, 3.94098499994524)
(36, 4.00524985637591)
(37, 4.08221201252174)
(38, 3.99515621210961)
(39, 4.10565479711281)
(40, 4.16848670350648)
(42, 4.19513170367148)
(43, 4.20491741741264)
(44, 4.22841198134678)
(45, 4.37575747282944)
(46, 4.36850659157323)
(48, 4.38484635090709)
(49, 4.29893296848091)
(50, 4.29747503008209)
(51, 4.35417111124355)
(52, 4.37504949167581)
(54, 4.48181898915407)
(55, 4.37271115919908)
(56, 4.45131056096132)
(57, 4.29655254017113)
(58, 4.4529942571222)
(60, 4.44871201676394)

};
\addlegendentry{Box Texture Log-Likelihood}

\addplot[smooth, color=magenta, densely dashed] plot coordinates {

(1,	0.538938646357673)
(2,	-1.2188903173404)
(3,	-2.01922176370573)
(4,	-2.49015931584942)
(6,	-2.8202197373803)
(7,	-3.061168661165)
(8,	-3.27000528662952)
(9,	-3.43783366739453)
(10, -3.61232767176272)
(12, -3.7481844970836)
(13, -3.87267657417563)
(14, -3.97064099738847)
(15, -4.07331279617044)
(16, -4.16391633754939)
(18, -4.279870807828)
(19, -4.33695470515768)
(20, -4.382)
(21, -4.47855430930408)
(22, -4.5499236417647)
(24, -4.63271565935505)
(25, -4.7035111455775)
(26, -4.75656955752207)
(27, -4.85058881038457)
(28, -4.88124115550103)
(30, -4.91796588660473)
(31, -4.97396488806502)
(32, -5.04333593121809)
(33, -5.10201999559924)
(34, -5.15850712173614)
(36, -5.22740577821115)
(37, -5.2625616794795)
(38, -5.28480364196929)
(39, -5.36411947041602)
(40, -5.36400189091317)
(42, -5.4416140750866)
(43, -5.45844664146651)
(44, -5.49006880100687)
(45, -5.52959614369407)
(46, -5.54436158184982)
(48, -5.59651945123625)
(49, -5.66051262528149)
(50, -5.67292017248733)

};
\addlegendentry{Bowl Uniform Training Loss}

\addplot[smooth, color=magenta] plot coordinates {

(1,	0.174117219496986)
(2,	1.37877975535626)
(3,	1.84981952706378)
(4,	2.13134021887939)
(6,	2.39579826213471)
(7,	2.61243398684393)
(8,	2.82525899569515)
(9,	2.79900004131713)
(10, 3.03418031415877)
(12, 3.13779624730565)
(13, 3.2405809464598)
(14, 3.18431978620521)
(15, 3.38290180449639)
(16, 3.34400646606171)
(18, 3.51790364081416)
(19, 3.37609354405642)
(20, 3.33585982904665)
(21, 3.22214313517934)
(22, 3.428293459886)
(24, 3.43722467101457)
(25, 3.23155588138075)
(26, 3.20592555007419)
(27, 3.25149766978556)
(28, 3.27147207868747)
(30, 3.37502548568648)
(31, 3.23907883779223)
(32, 3.10988619698789)
(33, 3.05420754301269)
(34, 3.14697612011648)
(36, 2.87540106459697)
(37, 2.85211888003565)
(38, 2.74154119020368)
(39, 2.64129761600896)
(40, 2.36830157882978)
(42, 2.13486388982944)
(43, 1.7928270092014)
(44, 2.09956525717419)
(45, 2.65425212967693)
(46, 1.86011159967125)
(48, 2.12175805436247)
(49, 1.57474966625966)
(50, 1.64889116608451)

};
\addlegendentry{Bowl Uniform Log-Likelihood}

\addplot[smooth, color=yellow, densely dashed] plot coordinates {

(1,	1.63221522112984)
(2,	0.868804144770352)
(3,	0.438112432795081)
(4,	0.143920277657719)
(6,	-0.082790940277167)
(7,	-0.306386116595559)
(8,	-0.529362012090078)
(9,	-0.730430668748137)
(10, -0.94540008954444)
(12, -1.14146646575548)
(13, -1.33772494543844)
(14, -1.45006379470303)
(15, -1.58574545413107)
(16, -1.70987204829259)
(18, -1.83191148084195)
(19, -1.94104274262243)
(20, -2.04399402669413)
(21, -2.13178122725653)
(22, -2.20525133135307)
(24, -2.2632922257357)
(25, -2.37449270753718)
(26, -2.46657355804348)
(27, -2.4930764335898)
(28, -2.55561724112402)
(30, -2.60162814043055)
(31, -2.65542122321342)
(32, -2.71806995963576)
(33, -2.77039869567055)
(34, -2.79789751086069)
(36, -2.84561996080389)
(37, -2.9104493014255)
(38, -2.86160666136006)
(39, -2.96739867611311)
(40, -3.01627267059402)
(42, -3.03641061640497)
(43, -3.09436959828903)
(44, -3.1119990064137)
(45, -3.19305085661399)
(46, -3.18801408561308)
(48, -3.24658878051226)
(49, -3.26915316676619)
(50, -3.31072919878794)

};
\addlegendentry{Can Uniform Training Loss}
\addplot[smooth, color=yellow] plot coordinates {

(1,	-1.28755084766363)
(2,	-0.647252117454735)
(3,	-0.293442563697235)
(4,	-0.084951334759632)
(6,	0.207500456959024)
(7,	0.397057651289352)
(8,	0.611591579159704)
(9,	0.856598133835398)
(10, 1.09027642919045)
(12, 1.26390013706151)
(13, 1.45104302156507)
(14, 1.64374687980617)
(15, 1.76410717682902)
(16, 1.85888766510605)
(18, 2.04867239660502)
(19, 2.09019688821271)
(20, 2.18110232649733)
(21, 2.30309707816521)
(22, 2.32021861718518)
(24, 2.45138476706622)
(25, 2.44757218363892)
(26, 2.58392094920281)
(27, 2.65735967975022)
(28, 2.67380913853075)
(30, 2.7567110821222)
(31, 2.77515890483444)
(32, 2.89000231617979)
(33, 2.79199699130644)
(34, 2.89483429583154)
(36, 2.8956138421265)
(37, 3.04659291411898)
(38, 2.98068370924443)
(39, 3.06727558778926)
(40, 3.1467121457825)
(42, 3.12238344482066)
(43, 3.21173959250094)
(44, 3.22506320216275)
(45, 3.26549650382106)
(46, 3.3108382869896)
(48, 3.22654375055585)
(49, 3.34856162962881)
(50, 3.28312992692234)

};
\addlegendentry{Can Uniform Log-Likelihood}

\addplot[smooth, color=pink, densely dashed] plot coordinates {

(1,	1.95452144430644)
(2,	1.40350560762396)
(3,	0.972512357863621)
(4,	0.639790386999424)
(6,	0.3535016669535)
(7,	0.111252994651892)
(8,	-0.168620672670022)
(9,	-0.416015972953234)
(10, -0.61533776607679)
(12, -0.87241262271629)
(13, -1.01298975877797)
(14, -1.18759761224339)
(15, -1.37036024367631)
(16, -1.47674411357339)
(18, -1.67250478949713)
(19, -1.79241165652204)
(20, -1.91258825591548)
(21, -2.08381605148315)
(22, -2.11749406329435)
(24, -2.21024727939966)
(25, -2.31962649650242)
(26, -2.37976319991534)
(27, -2.47256579216736)
(28, -2.53623242787461)
(30, -2.59508948065155)
(31, -2.67400942631622)
(32, -2.73071006281459)
(33, -2.82931850917304)
(34, -2.8621690332593)
(36, -2.90741969578302)
(37, -2.97074736113572)
(38, -2.92372405262136)
(39, -3.01575224079303)
(40, -3.11068212392911)
(42, -3.07613861323589)
(43, -3.13552020052772)
(44, -3.2526326161712)
(45, -3.23675939218322)
(46, -3.25650878806612)
(48, -3.2475971274115)
(49, -3.3049427780939)
(50, -3.28522465596745)

};
\addlegendentry{Box Uniform Training Loss}

\addplot[smooth, color=pink] plot coordinates {

(1,	-1.7596104525983)
(2,	-1.21777572148354)
(3,	-0.82628963750622)
(4,	-0.55519093282583)
(6,	-0.17212123263944)
(7,	-0.05043530163181)
(8,	0.390628998714188)
(9,	0.546106487527471)
(10, 0.81124049221673)
(12, 1.02470506850557)
(13, 1.31793608541608)
(14, 1.46233476453148)
(15, 1.79693564048627)
(16, 1.9202642357385)
(18, 2.14252522562883)
(19, 2.29059693496498)
(20, 2.41265212809206)
(21, 2.48417331078141)
(22, 2.6349398773737)
(24, 2.71057255469007)
(25, 2.79451458835302)
(26, 2.93223669651407)
(27, 3.05975100354536)
(28, 3.10984537687308)
(30, 3.22959010739966)
(31, 3.20537640821429)
(32, 3.41612383025601)
(33, 3.42270056433636)
(34, 3.48216972022353)
(36, 3.56830399332062)
(37, 3.41609499813778)
(38, 3.73411295842975)
(39, 3.75893317628898)
(40, 3.71452908498301)
(42, 3.67261503776661)
(43, 3.97937344464738)
(44, 3.91423732344775)
(45, 3.87441537533688)
(46, 3.96429829152077)
(48, 4.08406674630492)
(49, 3.96814861440957)
(50, 4.08702941884581)

};
\addlegendentry{Box Uniform Log-Likelihood}

\end{axis}
\end{tikzpicture}

}
          \caption{Training loss and log-likelihood on the validation set during the training process on the Uniform and Texture dataset. We early stop the training after 30 epochs when the log-likelihood starts stagnating.}
          \label{fig:training}
\end{figure}
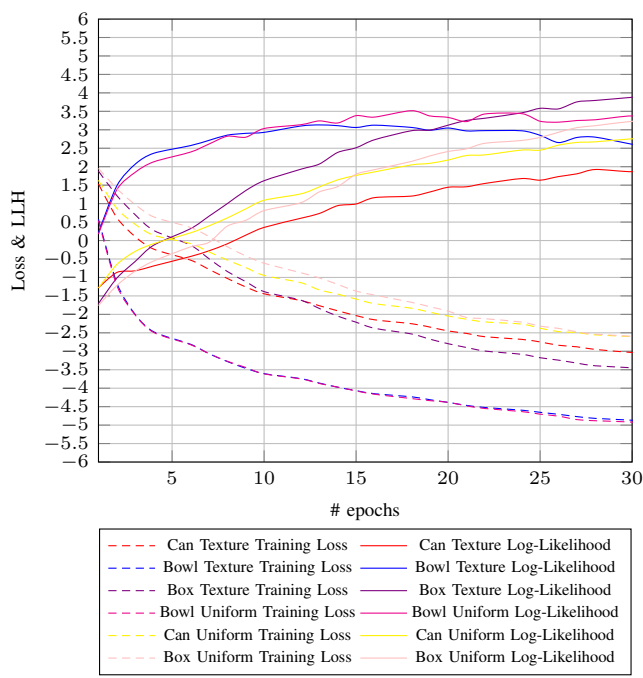

\begin{figure*}
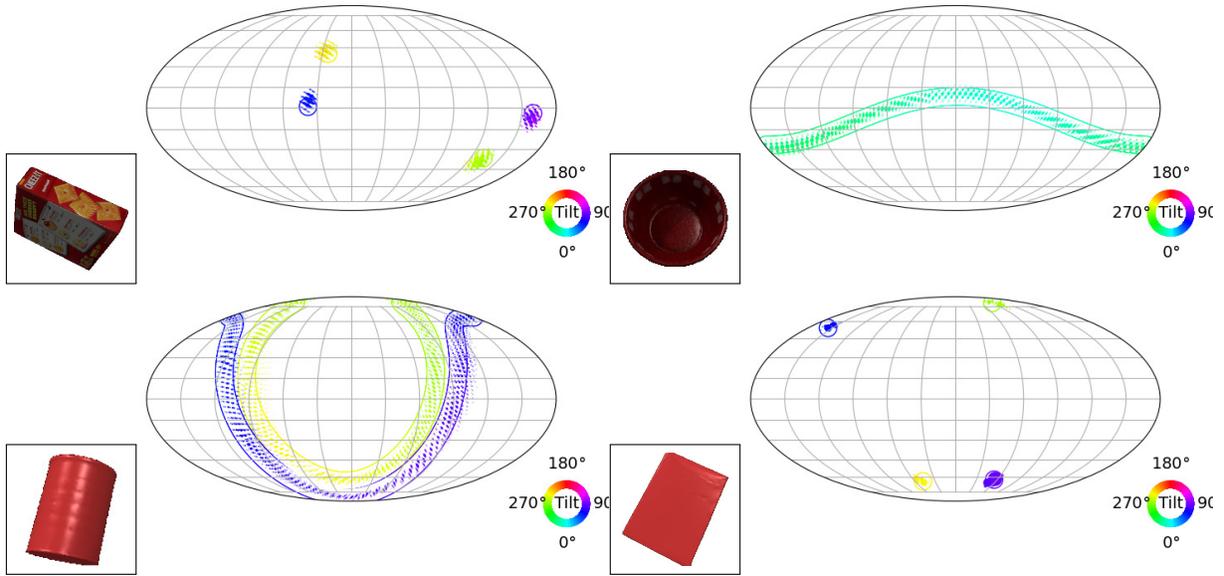

  \centering
 \setlength{\tabcolsep}{0.01cm}
 \newlength{\inimgw}\setlength{\inimgw}{1.5cm}
 \newlength{\vizimgw}\setlength{\vizimgw}{6.25cm}
\begin{tabular}{cccc}
    \framebox{\includegraphics[width=\inimgw, trim=20pt 20pt 20pt 20pt, clip]{figures/results/box/image_raw_0_6.png}} &
 \includegraphics[width=\vizimgw, trim=50pt 25pt 0pt 75pt, clip]{figures/results/box/visualization_rot_0_6.png} &
    \framebox{\includegraphics[width=\inimgw, trim=50pt 50pt 50pt 50pt, clip]{figures/results/bowl/image_0_6.png}} &
 \includegraphics[width=\vizimgw, trim=50pt 25pt 0pt 75pt, clip]{figures/results/bowl/visualization_rot_0_6.png} \\
    \framebox{\includegraphics[width=\inimgw, trim=50pt 50pt 50pt 50pt, clip]{figures/results/uniform_texture/image_0_1.png}} &
    \includegraphics[width=\vizimgw, trim=50pt 25pt 0pt 75pt, clip]{figures/results/uniform_texture/visualization_rot_0_1.png} &
    \framebox{\includegraphics[width=\inimgw, trim=10pt 10pt 10pt 10pt, clip]{figures/results/box/image_raw_0_6_uni.png}} &
 \includegraphics[width=\vizimgw, trim=50pt 25pt 0pt 75pt, clip]{figures/results/box/visualization_rot_0_6_uni.png} \\
\end{tabular}
    \caption{Orientation distributions predicted by our model. Top: Texture dataset. Bottom: Uniform dataset.
    The continuous lines and the circles represent the ground-truth symmetries and the dots represent the orientation hypotheses with a high estimated likelihood.
    The visualizations are generated using 294,912 orientation hypotheses ($S_i$=4).}
 \label{fig:results_viz}
\end{figure*}

\begin{figure*}
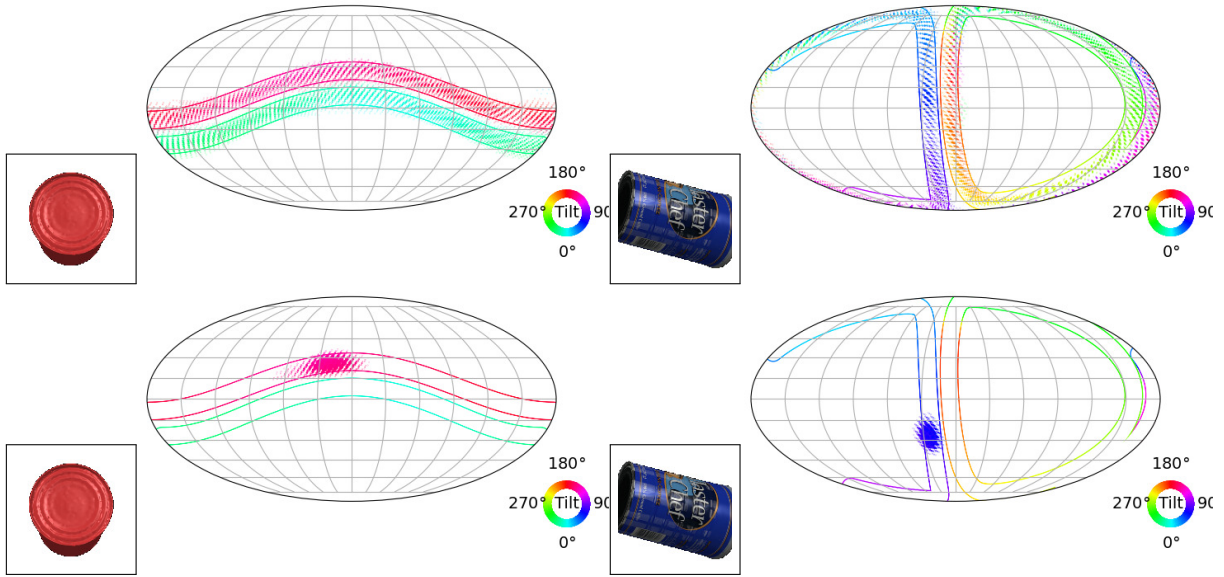

  \centering
 \setlength{\tabcolsep}{0.01cm}
 \setlength{\inimgw}{1.5cm}
 \setlength{\vizimgw}{6.25cm}
\begin{tabular}{cccc}
    \framebox{\includegraphics[width=\inimgw, trim=50pt 50pt 50pt 50pt, clip]{figures/results/uniform_texture/image_0_6.png}} &
    \includegraphics[width=\vizimgw, trim=50pt 25pt 0pt 75pt, clip]{figures/results/uniform_texture/visualization_rot_0_6.png} &
    \framebox{\includegraphics[width=\inimgw, trim=50pt 50pt 50pt 50pt, clip]{figures/results/material_texture_big/image_0_4.png}} &
 \includegraphics[width=\vizimgw, trim=50pt 25pt 0pt 75pt, clip]{figures/results/material_texture_big/visualization_rot_0_4.png} \\
    \framebox{\includegraphics[width=\inimgw, trim=50pt 50pt 50pt 50pt, clip]{figures/results/uniform_texture_gt/image_0_6.png}} &
    \includegraphics[width=\vizimgw, trim=50pt 25pt 0pt 75pt, clip]{figures/results/uniform_texture_gt/visualization_rot_0_6.png} &
    \framebox{\includegraphics[width=\inimgw, trim=50pt 50pt 50pt 50pt, clip]{figures/results/material_texture_big_gt/image_0_4.png}} &
 \includegraphics[width=\vizimgw, trim=50pt 25pt 0pt 75pt, clip]{figures/results/material_texture_big_gt/visualization_rot_0_4.png} \\
\end{tabular}
    \caption{Comparison of training with pseudo ground-truth generation and single ground-truth orientation on the Uniform (left) and Texture (right) dataset. 
    Top: Learning with pseudo ground-truth orientation. Bottom: Learning with a single ground-truth orientation. 
    The model trained using pseudo ground-truth annotations learns to complete orientation distribution, whereas the model trained using a single ground-truth
    learns only the single ground-truth but does not learn the complete orientation distribution.
    }
 \label{fig:results_comp}
\end{figure*}

\begin{figure*}
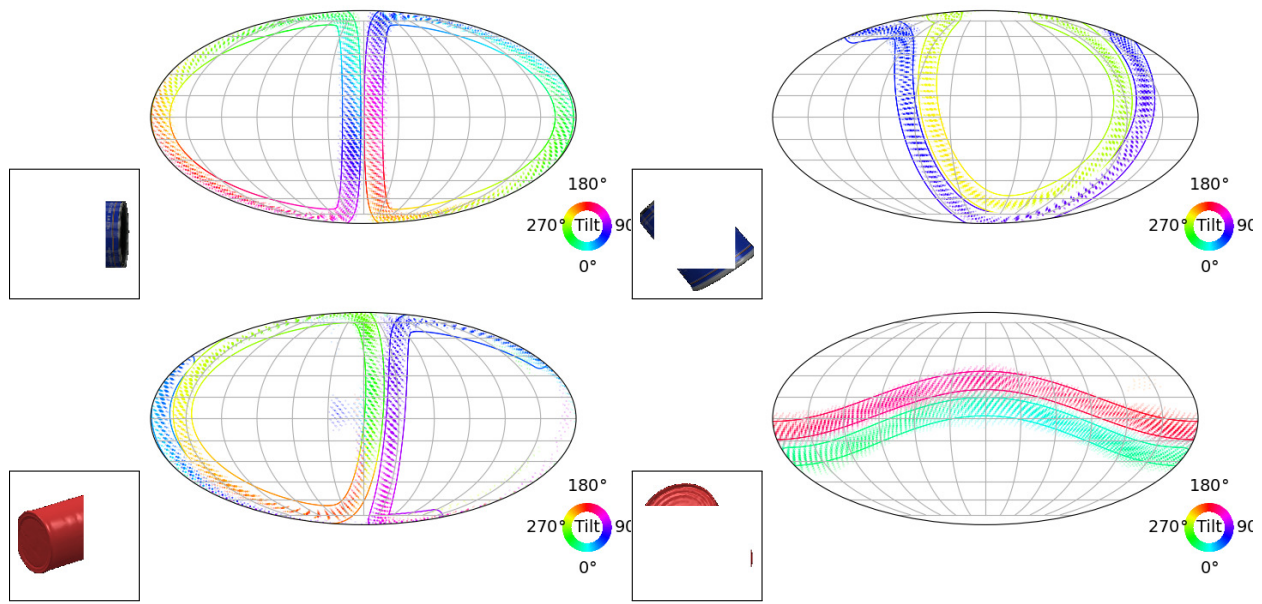

  \centering
 \setlength{\tabcolsep}{0.01cm}
 \setlength{\inimgw}{1.5cm}
 \setlength{\vizimgw}{6.5cm}
\begin{tabular}{cccc}
    \framebox{\includegraphics[width=\inimgw, trim=50pt 50pt 50pt 50pt, clip]{figures/results/material_texture_occlusion/image_0_6.png}} &
    \includegraphics[width=\vizimgw, trim=50pt 25pt 0pt 75pt, clip]{figures/results/material_texture_occlusion/visualization_rot_0_6.png} &
    \framebox{\includegraphics[width=\inimgw, trim=50pt 50pt 50pt 50pt, clip]{figures/results/material_texture_occlusion/image_0_12.png}} &
 \includegraphics[width=\vizimgw, trim=50pt 25pt 0pt 75pt, clip]{figures/results/material_texture_occlusion/visualization_rot_0_12.png} \\
    \framebox{\includegraphics[width=\inimgw, trim=50pt 50pt 50pt 50pt, clip]{figures/results/uniform_texture_occlusion/image_0_5.png}} &
    \includegraphics[width=\vizimgw, trim=50pt 25pt 0pt 75pt, clip]{figures/results/uniform_texture_occlusion/visualization_rot_0_5.png} &
    \framebox{\includegraphics[width=\inimgw, trim=50pt 50pt 50pt 50pt, clip]{figures/results/uniform_texture_occlusion/image_0_8.png}} &
    \includegraphics[width=\vizimgw, trim=50pt 25pt 0pt 75pt, clip]{figures/results/uniform_texture_occlusion/visualization_rot_0_8.png} \\
\end{tabular}
    \caption{Orientation distributions predicted by our model in the presence of occlusion. Top: Texture dataset. Bottom: Uniform dataset.
    }
 \label{fig:results_occ}
\end{figure*}

\begin{table}[h]
  \centering
  \footnotesize
  \setlength{\aboverulesep}{0pt}
  \setlength{\belowrulesep}{0pt}
    \caption{Results of models trained on different ground truths. 
    }
    \setlength\tabcolsep{4.6pt}
  \begin{tabular}{c|l|c|c|c|c|c|c}
    \toprule
      &GT & \multicolumn{3}{c|}{\thead{Without Occlusion}} & \multicolumn{3}{c}{\thead{With Occlusion}} \\ \cline{2-8}
      &&\thead{LLH}&\thead{MAAD}&\thead{Recall}&\thead{LLH}&\thead{MAAD}&\thead{Recall} \\
      &&&\thead{}&\thead{MAAD}& &\thead{}&\thead{MAAD} \\
      &&\textbf{$\uparrow$}&[°] \textbf{$\downarrow$}&[°] \textbf{$\downarrow$}&\textbf{$\uparrow$}&[°] \textbf{$\downarrow$}&[°] \textbf{$\downarrow$}\\
      
    \midrule
      \multirow{3}{*}{\STAB{\rotatebox[origin=c]{90}{can}}}
      &Single & 6.325  & 2.922  & 115.375 & 4.184  & 5.225 & 76.354\\
      &Analytical & 3.276  & 4.413 & 2.077 & 3.38 & 5.039 & 2.079 \\
      &Pseudo & 2.359  & 2.457 & 2.09 & 2.712 & 5.043 & 2.003 \\
    \midrule
    \multirow{3}{*}{\STAB{\rotatebox[origin=c]{90}{box}}}
      &Single & 7.096  & 2.651  & 72.586 &6.716 & 5.386 & 57.292 \\
      &Analytical & 3.656  & 3.444 & 1.978 & 3.436 & 4.955 & 1.955 \\
      &Pseudo & 4.452  & 7.481 & 2.31 & 4.544 & 8.845 &  2.353  \\
    \midrule
    \multirow{3}{*}{\STAB{\rotatebox[origin=c]{90}{bowl}}}
      &Single & 6.713  & 4.157  & 121.233 & 6.467 & 5.157 & 119.141 \\
      &Analytical & 4.39  & 8.534 & 2.839 & 3.835 & 9.266 & 2.223 \\
      &Pseudo & 4.544  & 8.845 & 2.353 &3.835 & 9.266 & 2.163 \\
      \midrule
      \multicolumn{2}{c|}{Pseudo Avg.}& 3.785  & 6.261 & 2.178 & 3.697 & 7.715 & 2.173\\
    \bottomrule
  \end{tabular}
  \vspace*{1mm}
  \\
  \textbf{$\uparrow$} indicates higher value better, whereas \textbf{$\downarrow$} indicates lower value better.
  \label{tab:gt_ablation}
\end{table}

We train our model for 50 epochs with 200 iterations per epoch. Each iteration consists of a batch of 64 images.
We assume that the object bounding box and the object segmentation mask are available to us.
Using the bounding box, we extract a crop of 560$\times$560 and resize it to the standard
ResNet input size 224$\times$224. Additionally, we mask the background pixels using the segmentation mask.
To compute $\mathcal{P}(\mathbf{R_{GT}} \given \mathcal{I})$ we use a grid $\{\mathbf{R^0}\}$ of cardinality 4,608 ($S_i$=2).
During testing, we compute the evaluation metrics using $\{\mathbf{R^n}\}$ of cardinality 294,912 ($S_i$=4).

Our model learns to predict the orientation on both Uniform and Texture datasets.
In~\cref{fig:training}, we show the training loss and the log-likelihood metric on the validation set during the training process.
We early stop the training when the log-likelihood metric starts to stagnate.
Qualitative samples of the predicted orientation distribution are shown in~\cref{fig:results_viz}.
From the visualizations, we observe that the model learns to predict the complete set of \textit{proper symmetries}.
In~\cref{tab:gt_ablation}, we report the validation scores. Overall, in the absence of occlusions, our model achieves a MAAD score of $\sim$6.2° and a Recall MAAD score of $\sim$2.2°.
A lower MAAD indicates high accuracy of the orientation predictions and a lower Recall MAAD shows that the model learns the complete set of \textit{proper symmetries}.

\begin{figure}[h]
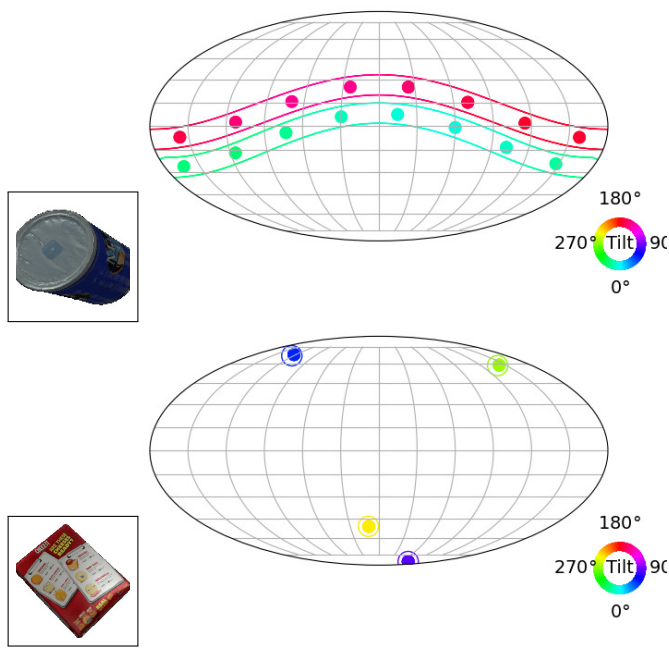

  \centering
 \setlength{\tabcolsep}{0.01cm}
 \newlength{\pgtimgw}\setlength{\pgtimgw}{2.5cm}
 \newlength{\pgtinimgt}\setlength{\pgtinimgt}{1.5cm}
 \newlength{\pgtvizimgt}\setlength{\pgtvizimgt}{7cm}
\begin{tabular}{cc}
    \framebox{\includegraphics[width=\pgtinimgt, trim=50pt 50pt 50pt 50pt, clip]{figures/results/pseudo_gt/can.png}} &
    \includegraphics[width=\pgtvizimgt, trim=50pt 25pt 0pt 75pt, clip]{figures/results/pseudo_gt/pgt_test_can.png} \\
    \framebox{\includegraphics[width=\pgtinimgt, trim=20pt 20pt 20pt 20ptt, clip]{figures/results/pseudo_gt/box.png}} &
 \includegraphics[width=\pgtvizimgt, trim=50pt 25pt 0pt 75pt, clip]{figures/results/pseudo_gt/pgt_test_box.png} \\
\end{tabular}
    \caption{Visualizing the generated pseudo ground-truth orientation labels. 
    Dots represent the generated pseudo ground-truth orientation labels, whereas the circles and the continuous lines represent the ground-truth orientation.
    The generated pseudo ground-truth orientation distribution correspond to the ground-truth orientation distribution with high accuracy.}
 \label{fig:pgt_comp}
\end{figure}

\begin{table}[h]
  \centering
  \footnotesize
  \setlength{\aboverulesep}{0pt}
  \setlength{\belowrulesep}{0pt}
  \caption{Evaluation of the pseudo ground-truth orientation labels}
  \begin{tabular}{c|c|c}
    \toprule
      Object & Dataset &\thead{MAAD}[°] \\
    \midrule
      \multirow{2}{*}{can}
      &\textit{Texture}& 1.44 \\
      &\textit{Uniform}& 3.12 \\

    \midrule
      \multirow{2}{*}{box}
      &\textit{Texture}&4.14 \\
      &\textit{Uniform}&4.3 \\
    \midrule
      \multirow{2}{*}{bowl}
      &\textit{Texture}&1.42\\
      &\textit{Uniform}&1.89\\
    \midrule
    \textbf{Average}&&2.72\\
    \bottomrule
  \end{tabular}
  \vspace*{1mm}

  \label{tab:pgt_comp}
\end{table}

\subsection{Occlusion}
Occlusion increases the complexity of computer vision problems. Pose estimation, in particular, is heavily impacted by the presence of occlusion.
To make our model robust against occlusion, we train our model by masking out random crops in the input images. We augment 80\% of images in each training batch randomly.
Between 10\% and 50\% of the image portions are occluded. In~\cref{fig:results_occ}, we present qualitative samples of the orientation predicted by our model in the presence of occlusion.
Despite the presence of occlusion, our model learns the orientation distribution well on both Texture and Uniform dataset. Our model performs only slightly worse compared to the occlusion-free model.
Quantitatively, in the presence of occlusion, our model achieves a MAAD score of $\sim$7.7° and a Recall MAAD score of $\sim$2.2°.
Interestingly, in terms of the Recall MAAD metric, the model trained using the single ground-truth
performs significantly better than in the case of no occlusions. This can be attributed to the uncertainty in the orientation estimate introduced by occlusion. 
Nevertheless, the model does not learn the correct orientation distribution from a single ground-truth pose.

\subsection{Comparison With Training Using Different Ground-Truths}
To evaluate the effectiveness of the pseudo ground-truth pose labeling scheme, 
we trained the implicit orientation estimation model using three different types of ground-truth poses:
single ground-truth pose used to render the image, complete set of \textit{proper symmetry} ground-truth poses generated analytically, and
the pseudo ground-truth poses generated using our pipeline.
The qualitative results are presented in~\cref{fig:results_comp}.
In~\cref{tab:gt_ablation}, we report the quantitative comparison results.
The single ground-truth model achieves a higher LLH score and a similar MAAD score for all objects, compared to both analytical and pseudo ground-truth models,
i.e. the single ground-truth model learns to estimate one single orientation precisely but fails to learn the symmetry orientations, whereas the other two models
manage to learn the complete set of symmetry orientations.
Moreover, the pseudo ground-truth model achieves results similar to the analytical model on all three metrics. 
Based on these results, we can conclude that the automatic pose labeling scheme is able generate pose labels with high accuracy 
and covers a sufficiently big portion of the set of \textit{proper symmetries} for the model to learn the symmetries.
Furthermore, as a measure of accuracy of the generated pseudo ground-truth orientation labels, we report in~\cref{tab:pgt_comp} the MAAD metrics of the generated pseudo ground-truth orientation labels for all three objects.
Overall, the average error rate of the generated pseudo ground-truth labels is $\sim$2.7°.
Among the three objects present in the dataset, the \textit{box} object has the highest MAAD error. This can be attributed to the fact
that \textit{box} exhibits discrete symmetry.
Generating pose labels for discrete symmetry is more difficult than for continuous symmetry.
As shown in~\cref{fig:pgt_comp}, the pseudo ground-truth pose labels correspond to the ground-truth orientation distribution with a high degree of accuracy.

\subsection{Backbone Ablations}
CNN models learn features that generalize well across datasets. However, the degree of generalization varies across different architectures.
In order to find the architecture best suited for usage as backbone feature extractor in the ImplicitPDF model,
we experimented with \mbox{ResNet}~\citep{he2016deep}, and \mbox{ConvNeXt}~\citep{liu2022conv} architectures.
Convolutional neural networks, in general, learn low-level image features at a high resolution in the initial layers and
high-level features at low resolution in the final layers~\citep{behnke2003hierarchical, SchulzB12, lecun2015deep}. 
For many computer vision tasks like object classification and object detection, high-level 
features---despite being low resolution---are ideal, whereas tasks like semantic segmentation benefit from access to low-level 
features~\citep{lin2017refinenet,schwarz2018rgb,ronneberger2015u}. To evaluate the effectiveness of different ResNet layers as feature extractors, we experimented with two
different ResNet backbone models---both derived from \mbox{ResNet-18} and taking images of size 224$\times$224 as input.
The first model is \mbox{ResNet-18} with only the last fully connected layer removed. 
The features extracted from this model form a vector of size 512. We call this model \mbox{ResNet-18-Full}.
The second model is ResNet-18 with the fully connected layer and the last convolutional block removed.
The features extracted from this model are of shape 1024$\times$14$\times$14. We call this model \mbox{ResNet-18-2}.
Additionally, we also experimented with the ConvNeXt Tiny model~\citep{liu2022conv}.
The features extracted from this model form a vector of size 768.
In~\cref{tab:resnet_comp}, we present the quantitative results of the comparison. In terms of both LLH and MAAD metrics on the Uniform and Texture datasets, all models perform
similarly. The size of the features extracted from the \mbox{ResNet-Full} backbone model is the smallest, though. Thus, we use the \mbox{ResNet-Full}
model as the backbone in the remaining experiments.

\begin{table}[h]
  \centering
  \footnotesize
  \setlength{\aboverulesep}{0pt}
  \setlength{\belowrulesep}{0pt}
  \caption{Comparison of different models as feature extractor.}
  \begin{tabular}{c|c|c|c|c}
    \toprule
      & \textit{Metric} & \multicolumn{2}{c|}{\thead{ResNet}} & \multicolumn{1}{c}{\thead{ConvNeXt}} \\ 
      \cline{3-4}
      &&\thead{18-Full}&\thead{18-2}& \\
      
    \midrule
      \multirow{2}{*}{\STAB{\rotatebox[origin=c]{90}{can}}}
      &LLH & 2.35  & 2.31  & 3.5 \\
      &MAAD& 2.45  & 3.16 & 3.46  \\
    \midrule
    \multirow{2}{*}{\STAB{\rotatebox[origin=c]{90}{box}}}
      &LLH & 4.45  & 4.23 & 4.64  \\
      &MAAD & 7.48  & 8.73 & 8.53  \\
    \midrule
    \multirow{2}{*}{\STAB{\rotatebox[origin=c]{90}{bowl}}}
      &LLH & 4.54  & 3.21 & 3.58\\
      &MAAD & 8.9  & 4.82 & 5.49\\

    \bottomrule
  \end{tabular}
  \vspace*{1mm}
  \label{tab:resnet_comp}
\end{table}

\subsection{Evaluation on T-Less Dataset}
\begin{figure}
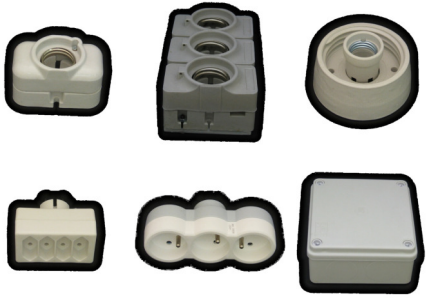

    \centering
     \setlength{\tabcolsep}{0.01cm}
     \setlength{\gtimgw}{2.cm}
    \begin{tabular}{ccc}
     \includegraphics[width=\gtimgw]{figures/tless/_obj_05.jpg} &
     \includegraphics[width=\gtimgw]{figures/tless/_obj_07.jpg} &
     \includegraphics[width=\gtimgw]{figures/tless/_obj_17.jpg} \\
     \includegraphics[width=\gtimgw]{figures/tless/_obj_20.jpg} &
     \includegraphics[width=\gtimgw]{figures/tless/obj_23.jpg} &
     \includegraphics[width=\gtimgw]{figures/tless/_obj_27.jpg} \\
    \end{tabular}
    \caption{T-Less dataset objects used for evaluating our method.}
    \label{fig:tless_objs}
\end{figure}

\begin{table}[h]
  \centering
  \footnotesize
  \setlength{\aboverulesep}{0pt}
  \setlength{\belowrulesep}{0pt}
  \caption{Comparison results on T-Less dataset.}
  \begin{tabular}{l|c|c}
    \toprule
     Method & \thead{LLH}&\thead{MAAD}[°] \\
    & \textbf{$\uparrow$} & \textbf{$\downarrow$}\\
    \midrule
    \citet{deng2022deep} & 5.3 & 23.1 \\
    \citet{gilitschenski2019deep}  & 6.9 & 3.4 \\
    \citet{prokudin2018deep}& 8.8 & 34.3 \\
    \citet{murphy2021implicit} & 9.8 & 4.1 \\ 
    Analytical & 5.7 & 1.7 \\
    Ours$^*$ & 5.23 & 3.6 \\
    
    \bottomrule
  \end{tabular}
  \vspace*{1mm}
  \\
  $^*$ Results from only a subset of the T-Less objects (shown in~\cref{fig:tless_objs}).
  \label{tab:tless}
\end{table}
The T-Less Dataset consists of RGB-D images of \mbox{texture-less} objects of varying sizes along with 6D pose annotations.
Training data consists of RGB-D images of individual objects placed in isolation with black background.
We evaluate our method on a subset of T-Less objects that exhibit geometric symmetries (shown in~\cref{fig:tless_objs}). We use the variant of the T-Less dataset proposed by~\citet{gilitschenski2019deep} in which
the training images provided in the original T-Less is split into training, validation, and test sets.
In~\cref{tab:tless}, we present quantitative results.
Our method achieves a MAAD score of $\sim$3.6° and an LLH score of 5.23. Since we report the metrics only for the objects that exhibit geometric symmetries,
an uncertainty always exists in terms of the object orientation. Thus, our model performs worse in terms of the LLH metrics, compared to other methods, 
but this does not affect the accuracy in terms of the MAAD metrics.
Moreover, compared to the model trained using analytically generated ground-truth orientation labels, which achieves a MAAD score of $\sim$1.7° and an LLH score of 5.7, our method
performs only slightly worse. This indicates that analytically generating ground-truth orientation labels for objects with complex geometric symmetry is not trivial. 
Moreover, the pose labeling scheme generates pseudo pose labels (SE(3)), whereas the analytical ground-truth generation is possible only for orientation labels (SO(3)). Having access to pose labels enables training complete pose estimation models.
Thus, the proposed automated pose labeling scheme serves as an efficient alternative to generating ground-truth orientation labels analytically.

\section{Discussion \& Conclusion}
In this article, we presented the ImplicitPDF model for learning object orientation for symmetrical objects.
We proposed a pseudo ground-truth labeling scheme to generate pose annotations
and used it to train the ImplicitPDF model without any manual pose annotations.
We quantified the advantages of multiple pseudo ground-truth labels over the single ground-truth pose
label for training the ImplicitPDF model
and the accuracy of the pose labeling scheme.
Moreover, by comparing with the models trained using analytically generated ground-truth orientation, we demonstrated the effectiveness of the automatic pose labeling scheme.
Overall, our method predicts the complete set of \textit{proper symmetries} for uniform color objects as well as for objects with texture with a high degree of accuracy.
In the future, we plan to
extend the implicit orientation model ($\mathbf{SO}(3)$) to 6D object pose ($\mathbf{SE}(3)$) to include the estimation of the translational component and reason about its uncertainty.

\section*{Acknowledgment}
This work has been funded by the German Ministry of Education and
Research (BMBF), grant no. 01IS21080, project Learn2Grasp: Learning
Human-like Interactive Grasping based on Visual and Haptic Feedback.

\printbibliography

\end{document}